\def\year{2019}\relax
\documentclass[letterpaper]{article} 
\usepackage{aaai19}  
\usepackage{times}  
\usepackage{helvet}  
\usepackage{courier}  
\usepackage{url}  
\usepackage{graphicx}  
\usepackage{subcaption} 
\usepackage{xcolor}
\usepackage{booktabs}
\usepackage{multirow}
\usepackage{amsmath}
\usepackage{amsfonts}
\usepackage{amsthm}
\usepackage{algorithm}
\usepackage[noend]{algpseudocode}

\graphicspath{{./}}

\newcommand{\comment}[1]{\textcolor{blue}{\# #1}}

\newcommand{\citet}[1]{\citeauthor{#1}~\shortcite{#1}}

\begin{document}
	%
    
\title{Switch-based Active Deep Dyna-Q: Efficient Adaptive Planning for Task-Completion Dialogue Policy Learning}

\author{Yuexin Wu$^\star$\quad Xiujun Li$^\dag$$^\ddag$\quad Jingjing Liu$^\dag$\quad Jianfeng Gao$^\dag$\quad Yiming Yang$^\star$\\
$^\star$Carnegie Mellon University\quad\quad $^\dag$Microsoft Research\\
$^\ddag$Paul G. Allen School of Computer Science \& Engineering, University of Washington\\
{\small \tt $^\star$\{yuexinw,yiming\}@cs.cmu.edu\quad\quad $^\dag$\{xiul,jingjl,jfgao\}@microsoft.com}
}

\maketitle

\begin{abstract}
Training task-completion dialogue agents with reinforcement learning usually requires a large number of real user experiences. The Dyna-Q algorithm extends Q-learning by integrating a world model, and thus can effectively boost training efficiency using simulated experiences generated by the world model. The effectiveness of Dyna-Q, however, depends on the quality of the world model - or implicitly, the pre-specified ratio of real vs. simulated experiences used for Q-learning. To this end, we extend the recently proposed Deep Dyna-Q (DDQ) framework by integrating a \emph{switcher} that automatically determines whether to use a real or simulated experience for Q-learning. Furthermore, we explore the use of active learning for improving sample efficiency, by encouraging the world model to generate simulated experiences in the state-action space where the agent has not (fully) explored. Our results show that by combining switcher and active learning, the new framework named as Switch-based Active Deep Dyna-Q (Switch-DDQ), leads to significant improvement over DDQ and Q-learning baselines in both simulation and human evaluations.\footnote{Source code is at https://github.com/CrickWu/Switch-DDQ.}
\end{abstract}
	
\section{Introduction}

Thanks to the increasing popularity of virtual assistants such as Apple's Siri and Microsoft's Cortana, there has been a growing interest in both industry and research community in developing task-completion dialogue systems \cite{gaosurvey}. Dialogue policies in task-completion dialogue agents, which control how agents respond to user input, are typically trained in a reinforcement learning (RL) setting~\cite{DBLP:journals/pieee/YoungGTW13,levin1997learning}. RL, however, usually requires collecting experiences via direct interaction with real users, which is a costly data acquisition procedure, as real user experiences are much more expensive to obtain in the dialogue setting than that in simulation-based game settings (such as Go or Atari games)~\cite{DBLP:journals/nature/MnihKSRVBGRFOPB15,DBLP:journals/nature/SilverHMGSDSAPL16}.

One common strategy is to train policies with user simulators that are developed from pre-collected human-human conversational data \cite{schatzmann2007agenda,li2016user}. Dialogue agents interacting with such user simulators do not incur any real-world cost, and can in theory generate unlimited amount of simulated experiences for policy training. The learned policy can then be further fine-tuned using small amount of real user experiences~\cite{dhingra2016towards,su2016continuously,lipton2016efficient,li2017end}.

Although simulated users provide an affordable alternative, they may not be a sufficiently truthful approximation to human users. The discrepancy between simulated and real experiences inevitably leads to strong bias. In addition, it is very challenging to develop a high-quality user simulator, because there is no widely accepted metric to assess the quality of user simulators \cite{pietquin2013survey}. It remains a controversial research topic whether training agents through user simulators is an effective solution to building dialogue systems.

\begin{figure*}[t!]
	\centering
	\begin{subfigure}[t]{0.5\textwidth}
		\centering
		\includegraphics[width=0.7\textwidth]{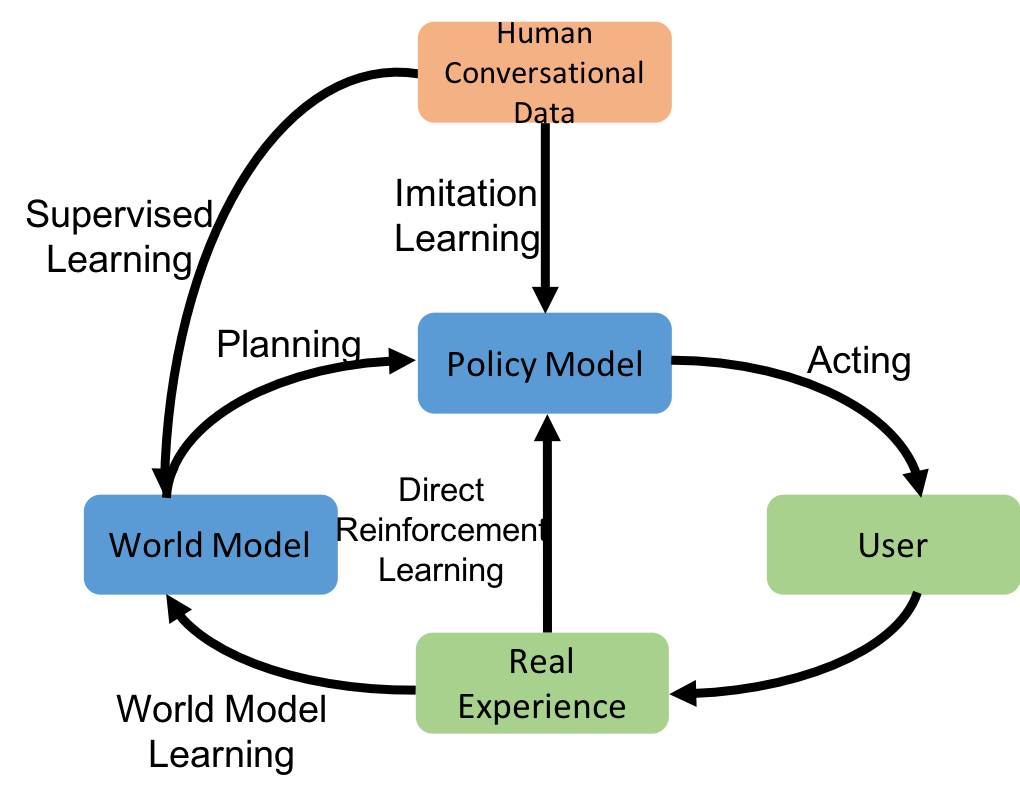}
		\caption{DDQ framework}
		\label{fig:ddq}
	\end{subfigure}%
	~ 
	\begin{subfigure}[t]{0.5\textwidth}
		\centering
		\includegraphics[width=0.7\textwidth]{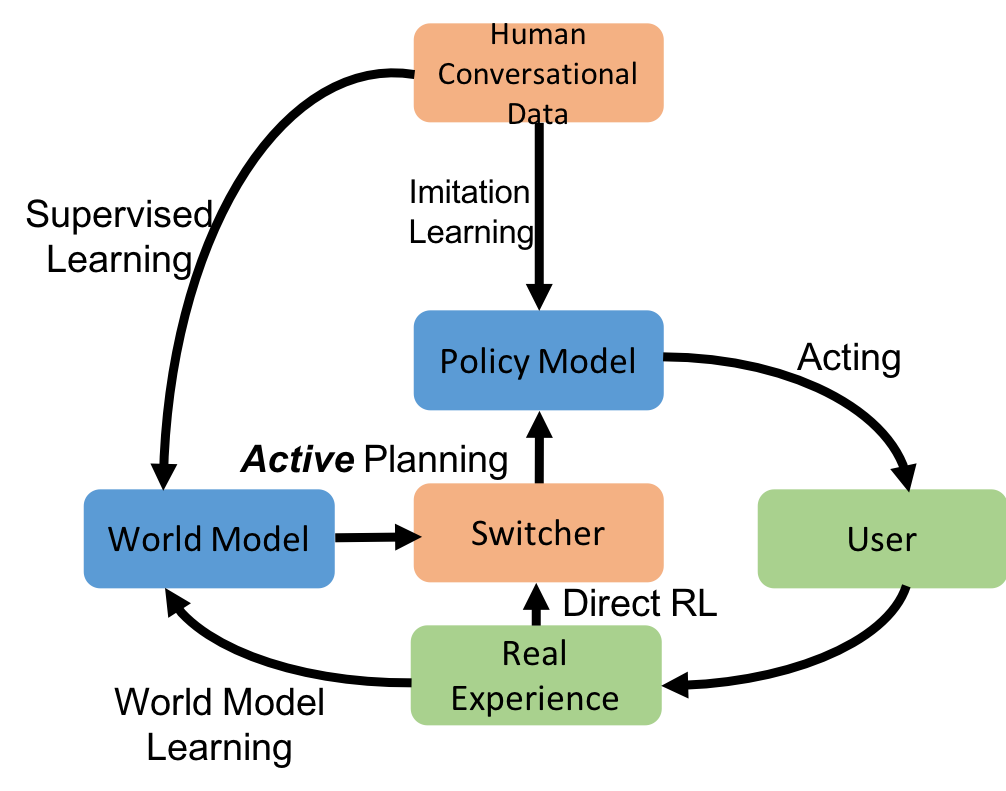}
		\caption{Proposed Switch-DDQ framework}
		\label{fig:saddq}
	\end{subfigure}
	\caption{Designs of RL agents for dialogue policy learning in task-completion dialogue systems}
    \vspace{-0.1in}
\end{figure*}

Recently, \citet{Peng2018DeepDynaQ} proposed Deep Dyna-Q (DDQ), an extension of the Dyna-Q framework \cite{sutton1990integrated}, which integrates planning into RL for task-completion dialogue policy learning.
As illustrated in Figure~\ref{fig:ddq}, DDQ incorporates a trainable user simulator, referred to as the \textit{world model}, which can mimic real user behaviors and generate simulated experience. The policy of the dialogue agent can be improved through either (1) real user experiences via \emph{direct RL}; or (2) simulated experiences via \emph{indirect RL} or \emph{planning}.

\begin{figure}[ht!] 
\centering
\includegraphics[width=0.9\linewidth]{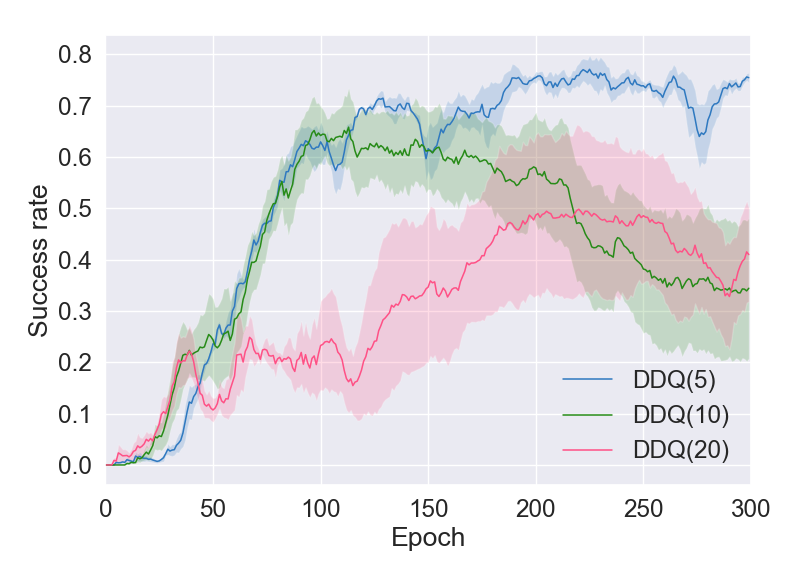}
\caption{The learning curves of DDQ($K$) without heuristics where $(K-1)$ denotes the number of planning steps. The curves are sensitive to $K$ values and may deteriorate in the later phase due to the low-quality simulated experiences.}
\label{fig:DDQ_perf}
\end{figure}

DDQ is proved to be sample-efficient in that a reasonable policy can be obtained using a small number of real experiences, an affordable training process thanks to the integration of planning into RL.
However, the effectiveness of DDQ depends, to a large degree, upon the way we control the ratio of real vs. simulated experiences used in different stages of training. For example, \citet{Peng2018DeepDynaQ} pointed out that although aggressive planning (i.e., policy learning using a large number of simulated experiences) often helps improve the performance in the beginning stage of training when the agent is not sensitive to the low-quality experiences, such aggressive planning might hurt the performance in the later stage when the agent is more susceptible to noise, as illustrated in Figure~\ref{fig:DDQ_perf}. Carefully designed heuristics are essential to set the ratio properly. For example, we might decrease the number of simulated experiences during the course of training. However, such heuristics can vary with different settings, and thus significantly limits the wide application of DDQ in developing real-world dialogue agents.

Another limitation of DDQ is that the world model generates simulated experiences by uniformly sampling user goals. However, training samples in the state-action space unexplored or less explored by the dialogue agent are usually more desirable in order to avoid bias. This is the problem that many active learning methods try to address. In this paper, we present a new variant of DDQ that addresses these two issues.

Our method is inspired by the recent study of \citet{su2018discriminative}, which tries to balance the use of simulated and real experience by measuring the quality of simulated experiences using a machine-learned \emph{discriminator}. The more simulated experiences are used if their quality is higher. Their approach
demonstrates some limited success, and suffers from two shortcomings. First, it does not take into account the fact that the agent in different training stages might require simulated experiences of different qualities. Second, it still uniformly samples user goals and is not as sample-efficient as it should be (e.g., by using active learning). 


In this paper, we propose a new framework, called Switch-based Active Deep Dyna-Q (Switch-DDQ), to significantly improve DDQ's sample efficiency.
As illustrated in Figure~\ref{fig:saddq}, we incorporate a \textit{switcher} to automatically determine whether to use real or simulated experiences at different stages of dialogue training, eliminating the dependency on heuristics. The switcher is implemented based on an LSTM model, and is jointly trained with the dialogue policy and the world model. 
Moreover, instead of randomly sampling simulated experiences, the world model adopts an \textit{active} sampling strategy that generates simulated experiences from the state-action space that has not been (fully) explored by the dialogue agent. Experiments show that this active sampling strategy can achieve a performance that is comparable to the original DDQ method but by using a much smaller amount of real experiences.


The work present in this paper contributes to the growing family of model-based RL methods, and can potentially be applied to other RL problems. To the best of our knowledge, Switch-DDQ is the first learning framework that conducts active learning in a task-completion dialogue setting. The contributions of this work are two-fold:
\begin{itemize}
\item We propose a Switch-based Active Deep Dyna-Q framework to incorporate active learning into the Dyna-Q framework for dialogue policy learning, providing a mechanism of automatically balancing the use of simulated and real user experiences.
\item We validate the superior performance of Switch-DDQ by building dialogue agents for the movie-ticket booking task. The effectiveness of active learning and switcher is verified by simulation and human evaluations. 
\end{itemize}

\section{Model Architecture}
\begin{figure}[htbp]
\centering
\includegraphics[width=1\linewidth]{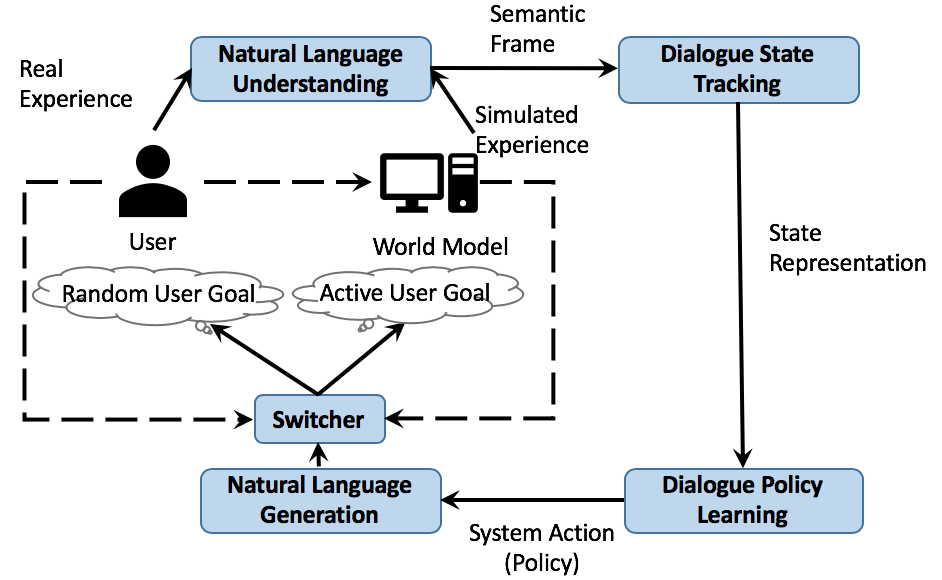}
\caption{Switch-DDQ for dialogue policy learning.}
\label{fig:paradigm}
\end{figure}

We depict our Switch-DDQ pipeline in Figure~\ref{fig:paradigm}. The agent consists of six modules: (1) an LSTM-based natural language understanding (NLU) module \cite{hakkani2016multi} for extracting user intents/goals and calculated their associated slots; (2) a state tracker \cite{mrkvsic2016neural} for tracking dialogue states; (3) a dialogue policy that makes choice of the next action by using the information of the current dialogue state; (4) a model-based natural language generation (NLG) module which outputs natural language responses \cite{wen2015semantically}; (5) a world model for generating simulated user actions and simulated rewards based on active user goal selection; and (6) an RNN-based switcher for selecting the source of data (simulated or real experiences) for dialogue policy training. The solid lines in the figure illustrate the iterative dialogue policy training loop, while the dashed lines show the flow of data in training the world model and switcher. 

The optimization of Switch-DDQ comprises four steps: (1) \textit{direct reinforcement learning}: the agent conducts direct interactions with a real user, where the generated real experiences are directly used to improve the dialogue policy; (2) \textit{active planning}: the agent interacts with the simulator and improves the policy using the simulated experiences; (3) \textit{world model learning}: the world model receives real experiences and updates itself; and (4) \textit{switcher training}: the switcher is learned and refined using both real and simulated experiences. Each step is detailed in the subsections below. The iterative Switch-DDQ algorithm, described in pseudo-code, is shown in Algorithm~\ref{alg:saddq}.

\subsection{Direct Reinforcement Learning and Planning}
Typically, dialogue policy learning can be formulated as a Markov Decision Process in the RL setting, a task-completion dialogue could be viewed as a sequence of (state, action, reward) tuples. We employ the Deep Q-network (DQN)~\cite{DBLP:journals/nature/MnihKSRVBGRFOPB15} for training the dialogue policy (line \ref{alg:trn:agent} in Algorithm~\ref{alg:saddq}). Both the direct reinforcement learning and planning are accomplished using the same Q-learning algorithm using the simulated and real experiences, respectively.

Specifically, at each step, the agent receives the state $s$ and selects an action $a$ to carry into the next dialogue turn. The action $a$ is chosen using the exploration policy based on $\epsilon$-greedy, where there is probability $\epsilon$ a random action being executed or otherwise the action that maximizes the $Q(s,a;\theta_Q)$ function. The function $Q(\cdot)$ is parameterized by a Multi-Layer Perceptron (MLP) parameterized by $\theta_Q$. 
Afterwards, the agent observes a reward $r$ from the environment, 
and a corresponding response $a^u$ from either a real user or the simulator, updating the dialogue state to $s'$ until reaching the end of a dialogue. The experience $(s,a,r,a^u,s')$ is then stored into the user experience buffer $B^u$ or simulator experience buffer $B^s$ respectively. Function $Q(\cdot)$ can be improved using experiences stored in the buffers.

In the implementation, we optimize the parameter $\theta_Q$ w.r.t. the mean-squared loss:
\begin{align}
\mathcal{L}(\theta_Q)&=\mathbb{E}_{(s,a,r,s')\sim B^s\cup B^u}\left[\left(y-Q(s,a;\theta_Q)\right)\right]\\
y&=r+\gamma\max_{a'} Q'(s',a';\theta_{Q'})
\end{align}
where $Q'(\cdot)$ is a copy of the previous version of $Q(\cdot)$ and is only updated periodically and $\gamma\in[0,1]$ is the discount factor. $Q(\cdot)$ is updated using back-propagation and mini-batch gradient descent.

\begin{algorithm}[htbp]
\caption{Switch-based Active Deep Dyna-Q}
\label{alg:saddq}
	\begin{algorithmic}[1]
		\Procedure{Switch-DDQ TrainingPipeline}{}
		\For {$i\gets 1:\text{max\_epoch}$} \label{alg:trn:start_for}
		\State\textit{user} randomly picks a user goal $g^u$ \label{alg:trn:user_pick}
		\State Generate real experience $e^u$ from \textit{user} based on $g^u$ into $B^u$ \label{alg:trn:store_real}
		\Repeat \label{alg:trn:start_repeat}
		\State \textit{Actively} select a user goal $g^s$ based on the validation results \comment{see Algorithm~\ref{alg:active}}
		 \label{alg:trn:active}
		\State Generate simulated experience $e^s$ from \textit{simulator} based on $g^s$ into $B^s$ \label{alg:trn:store_fake}
		\State Evaluate \textit{quality} of $e^s$ through \textit{switcher} \label{alg:trn:switcher}
		\Until{ \textit{quality} $<$ threshold} \label{alg:trn:end_repeat} 
		\State Train \textit{simulator} on $B^u$
		\State Train \textit{switcher} on $B^u,B^s$
		\State Train \textit{agent} on $B^u, B^s$\label{alg:trn:agent}
		\State Evaluate \textit{simulator} on validation set \label{alg:trn:validation}
		\EndFor
		\EndProcedure		
	\end{algorithmic}
\end{algorithm}

\subsection{Active Planning based on World Model}
In a typical task-completion dialogue \cite{schatzmann2007agenda}, a user begins a conversation with a particular goal in mind $G$ which consists of multiple constraints. For example, in the movie-ticket-booking scenario, the constraints can be the place of the theater, the number of tickets to buy, and the name of the movie. An example of a user goal is \texttt{request(theater;numberofpeople=2, moviename=mission\_impossible)}, which is presented in its natural language form as ``\texttt{in which theater can I buy two tickets for mission impossible}''. Although there is no explicit restriction for the range of user goals in real experiences, in the stage of planning, the world model can \emph{selectively} generate the simulated experiences in the state-action space that are not (fully) explored by the dialogue agent, based on a specific set of user goals, to improve sample efficiency. We call our planning \emph{active planning} because it is a form of active learning.

The world model for active planning consists of two parts: (1) a user goal sampling module that samples a proper user goal at the start of a dialogue; (2) a response generation module that imitates real users' interaction with the agent to generate for each dialogue turn the user action, reward and the user's decision whether to terminate the dialogue.

\begin{itemize}
\item \textit{Active user goal sampling module}. Assume that we have collected large amounts of user goals from human-human conversational data. These user goals can be grouped into different categories, each with different constraints, amounting to different scales of difficulties. The key observation is that, during the training process, while monitoring the performance of the agent policy on validation set, we can gather detailed information about the impact of each category of user goals on the performance improvement of the dialogue agent e.g., in terms of the success rate (line~\ref{alg:trn:validation} in Algorithm~\ref{alg:saddq}). The detailed information can be used to measure 
the cost (or gain) in the active learning setting \cite{russo2018tutorial,auer2002finite} and guide the world model how to sample user goals. 
	
Suppose there are $k$ different categories of user goals. At each epoch, the failure rate of each category estimated on the validation set is denoted as $f_i$ and the number of samples for the estimation is $n_i$. For simplicity, denote the summation of $n_i$ as $N=\sum_i n_i$. Then, the active sampling routine (line~\ref{alg:trn:active} in Algorithm~\ref{alg:saddq}) can be expanded as 
	\begin{algorithm}[htbp]
		\caption{Active Sampling Routine}	\label{alg:active}
		\begin{algorithmic}[1]
			\Procedure{Active User Goal Sampling}{}
			\State Draw a number $p_i$ for each category following $p_i\sim\mathcal{N}\left(f_i, \sqrt{\frac{k\ln N}{n_i}}\right)$ \label{alg:active:draw}
			\State Select the user goal $i$ with the maximum $p_i$ value
			\EndProcedure		
		\end{algorithmic}
	\end{algorithm}
	 
Here, $\mathcal{N}$ is the Gaussian distribution for introducing randomness. The Thompson-Sampling-like \cite{russo2018tutorial} sub-routine of Algorithm~\ref{alg:active} is motivated by two observations: (1) on average, categories with larger failure rate $f_i$ are more preferable as they inject more difficult cases (containing more useful information to be learned) based on the current performance of the agent policy. The generated data (simulated experiences) are generally associated with the steepest learning direction and can prospectively boost the training speed; (2) categories that are estimated less reliably (due to a smaller value of $n_i$ value) may have a large de facto failure rate, thus worth being allocated with more training instances to reduce the uncertainty. $\sqrt{\frac{k\ln N}{n_i}}$ is the measurement of the uncertainty of $f_i$, serving the role of variance in the Gaussian. Thus, the categories with high uncertainty are still likely to be selected even if the failure rate is small. 
	
\item \textit{Response generation module}. We utilize the same design of the world model in \citet{Peng2018DeepDynaQ}. Specifically, we parameterize it using a multi-task deep neural network \cite{liu2015representation}. Each time the world model observes the dialogue state $s$ and the last action from the agent $a$, it passes the input pair $(s,a)$ through an MLP $M(s,a;\theta_M)$ generating a user action $a^u$, a regressed reward $r$ and a binary terminating indicator signal $t$. The MLP has a common sharing representation in the first layer (referred to as layer $h$). The computation for each term can be shown as below: 
\begin{align}
	h&=\tanh(W_h(s,a)+b_h)\\
	a^u&=\texttt{softmax}(W_ah+b_a)\\
	r&=\tanh(W_rh+b_r)\\
	t&=\texttt{sigmoid}(W_th+b_t)
	\end{align}
\end{itemize}

\subsection{Switcher}
At every step of training, the switcher needs to decide whether the dialogue agent should be trained using simulated or real experience 
(lines~\ref{alg:trn:switcher}-\ref{alg:trn:end_repeat} in Algorithm~\ref{alg:saddq}).

The switcher is based on a binary classifier implemented using a Long Short-Term Memory (LSTM) model \cite{hochreiter1997long}. Assume that a dialogue is represented as a sequence of dialogue turns, denoted by $\{(s_i, a_i, r_i)\}$, $i=1,...,N$, where $N$ is the number of dialogue turns of the dialogue. Q-learning takes a tuple in the form of $(s,a,r,s')$ as a training sample, which can be extracted from two consecutive dialogue turns in a dialogue. Now, the design choice of switcher is whether the classifier is turn-based or dialogue-based. We choose the former, though a bit anti-intuitive, for data efficiency. There is an order of magnitude larger number of turns than that of dialogues. As a result, a turn-based classifier can be more reliably trained than a dialogue-based one. Then, given a dialogue, we score the quality of each of its dialogue turns, and then averages these scores to measure the quality of the dialogue (line~\ref{alg:trn:active} in Algorithm~\ref{alg:saddq}). If the dialogue-level score is below a certain threshold, the agent switches to interact with real users.
 


Note that each dialogue turn is scored by taking into account its previous turns in the same dialogue. Given a dialogue turn $(s_t,a_t,r_t)$ and its history $h=((s_1,a_1,r_1),(s_2,a_2,r_2),...,(s_{t-1},a_{t-1},r_{t-1}))$ We use LSTM to encode $h$ using the hidden state vector, and output a turn-level quality score via a sigmoid layer:
\begin{align}
\text{Score}((s,a,r),h;\theta) = \texttt{sigmoid}( \text{LSTM}((s,a,r),h; \theta))
\end{align}


Since we store user experiences and simulated experiences in the buffers $B^u$ and $B^s$, respectively (lines~\ref{alg:trn:store_real}, \ref{alg:trn:store_fake} in Algorithm~\ref{alg:saddq}), the training of $\text{Score(.)}$ follows a similar process of minimizing the cross-entropy loss as in the common domain adversarial training setting \cite{ganin2016domain} using mini-batches:
\begin{align}
	&\min_{\theta_S} \mathbb{E}_{(s,a,r),h\sim B^u}\log \left( \text{Score} \left( (s,a,r),h;\theta \right) \right) \nonumber\\
	&+\mathbb{E}_{(s,a,r),h\sim B^s}\log \left( 1- \text{Score}((s,a,r),h;\theta) \right)
\end{align}

Since the experiences stored in $B^s$ and $B^u$ change during the course of dialogue training, the score function of the switcher is updated accordingly, thus automatically adjusting how much planning to perform at different stages of training. 


\section{Experiments}
We evaluate the proposed Switch-DDQ framework in the movie-ticket booking domain, in two settings: simulation and human evaluation. 

\subsection{Dataset}
For experiments, we use a movie-ticket booking dataset which contains raw conversational data collected via Amazon Mechanical Turk. The dataset is manually labeled based on a schema defined by domain experts. As shown in Table \ref{tab:annotations}, the annotation schema consists of 11 intents and 16 slots. In total, the dataset contains 280 labeled dialogues, the average length of which is 11 turns. 

\begin{table}[htbp]
\centering
\begin{tabular}{cllll}
\toprule
\multicolumn{5}{c}{Annotations} \\
\midrule
\multirow{3}[2]{*}{Intent} & \multicolumn{4}{l}{request, inform, deny, confirm\_question,} \\
& \multicolumn{4}{l}{confirm\_answer, greeting, closing, not\_sure,} \\
& \multicolumn{4}{l}{multiple\_choice, thanks, welcome} \\
\midrule
\multirow{4}[2]{*}{Slot} & \multicolumn{4}{l}{city, closing, date, distanceconstraints,} \\
& \multicolumn{4}{l}{greeting, moviename, numberofpeople,} \\
& \multicolumn{4}{l}{price, starttime, state, taskcomplete, theater,} \\
& \multicolumn{4}{l}{theater\_chain, ticket, video\_format, zip} \\
\bottomrule
\end{tabular}%
\caption{The data annotation schema}
\label{tab:annotations}%
\end{table}%

\subsection{Baselines}
We compare the effectiveness of the Switch-DDQ agent with several baselines:
\begin{itemize}
\item \textbf{DQN} agent is implemented with only direct reinforcement learning in each training epoch (without lines~\ref{alg:trn:start_repeat}-\ref{alg:trn:end_repeat} in Algorithm~\ref{alg:saddq}).
\item The \textbf{DQN($K$)} has $(K-1)$ times more real experiences than the DQN agent (repeat lines~\ref{alg:trn:user_pick}-\ref{alg:trn:store_real} in Algorithm~\ref{alg:saddq} $K$ times). The performance of DQN($K$) can be viewed as the upper bound of DDQ ($K$), with the same number of planning steps $(K - 1)$ (they have the same training setting and the same amount of training samples during the entire learning process).
\item The \textbf{DDQ($K$)} agents are learned using a jointly-trained world model initiated from human conversational data, with $(K -1)$ planning steps (replace lines~\ref{alg:trn:start_repeat}-\ref{alg:trn:end_repeat} in Algorithm~\ref{alg:saddq} with a $(K-1)$-round loop).
\item 
The proposed \textbf{Switch-DDQ} agents are updated as described in Algorithm~\ref{alg:saddq}. Note that there is no parameter $K$ in the agent, as real/simulated ratio is automatically controlled by the switcher module.
\end{itemize}

\subsection{Implementation Details}
\subsubsection{Agent and Hyper-parameter Settings}
We use an MLP to parameterize function $Q(\cdot)$ in all the agent variants (DQN, DDQ and Switch-DDQ). The MLP has one hidden layer of $80$ neurons with ReLU \cite{nair2010rectified} activation function. The $\epsilon$-greedy policy is adopted to explore the action space. The discount factor $\gamma$ for future rewards is set to $0.9$. For DDQ($K$), as the number of real and simulated experiences is different at each epoch, the buffer sizes of $B^u$ and $B^s$ are generally set to $2000$ and $2000 \times K$, respectively. For Switch-DDQ, we observed that the results are not sensitive to the buffer size of $B^s$, so we set it to $2000\times 5$ for all settings. 

We randomly initialize the parameters in all neural networks and empty both experience buffers $B^u$ and $B^s$ in the beginning. The RMSProp \cite{hinton2012rmsprop} algorithm is used to perform optimization over all the parameters where 
the learning rate is set to $0.001$. We also apply the gradient clipping trick to all parameters with a maximum norm of $1$ to prevent possible gradient explosion issues. At the beginning of each epoch (line~\ref{alg:trn:start_for} in Algorithm~\ref{alg:saddq}), the reference copy $Q'(\cdot)$ is updated. Each simulated dialogue contains less than 40 turns. Conversations exceeding the maximum number of turns are counted as failed. In order to train the agents more efficiently, we utilized the imitation learning method called Reply Buffer Spiking (RBS) \cite{lipton2016efficient} at the initial stage to build a simple rule-based agent trained from human conversational data. The trained agent is then used to pre-fill the real experience replay buffer $B^u$ with a total of $50$ complete dialogues before training all the variants of the agent. 

\subsubsection{World Model}
We employ an MLP world model for DDQ and Switch-DDQ. The shared hidden layer is set to have size $160$ with hyperbolic tangent activation. The state and action input are encoded through a linear layer of size $80$. We pre-fill each $n_i$ as $5$ to prevent division by $0$ error, during the calculation of the Gaussian variance (line~\ref{alg:active:draw} in Algorithm~\ref{alg:active}).

\subsubsection{Switcher}
The LSTM switcher has a hidden layer with $126$ cells. Similar to the world model, states and actions are passed through a linear layer of size $80$ as inputs at each time step. The switcher adopts an annealing threshold w.r.t. the epoch number to decide the quality of each dialogue turn. If the average dialogue episode score passes a certain threshold, all the high-quality predictions are pushed into buffer $B^s$.\footnote{See the code for specific hyper-parameters.} 

\begin{table*}[ht!]
\centering
\begin{tabular}{lrrrrrrrrr}
\\ \hline
\multirow{2}{*}{Agent}& \multicolumn{3}{c}{Epoch = 100} & \multicolumn{3}{c}{Epoch = 200} & \multicolumn{3}{c}{Epoch = 300} \\ 
\cline{2-10}
 & Success & Reward & Turns & Success & Reward & Turns & Success & Reward & Turns \\ 
\midrule
DQN   & 0.2867 & -17.35 & 25.51 & 0.6733 & 32.48 & 18.64 & 0.7667 & 46.87 & 12.27 \\
DQN(5) & \textit{0.7667} & \textit{46.74} & \textit{12.52} & \textit{0.7867} & \textit{49.46} & \textit{11.88} & \textit{0.8000} & \textit{50.81} & \textit{12.37} \\
DDQ(5) & 0.6200 & 25.42 & 19.96 & 0.7733 & 45.45 & 16.69 & 0.7467 & 43.22 & 14.76 \\
DDQ(10) & \textbf{\textcolor{blue}{0.6800}} & 34.42 & 16.36 & 0.6000 & 24.20 & 17.60 & 0.3733 & -2.11 & 15.81 \\
DDQ(20) & 0.3333 & -13.88 & 29.76 & 0.4467 & 5.39  & 18.41 & 0.3800 & -1.75 & 16.69 \\
Switch-DDQ & 0.5200 & 15.48 & 15.84 & \textbf{\textcolor{blue}{0.8533}} & 56.63 & 13.53 & \textbf{\textcolor{blue}{0.7800}} & 48.49 & 12.21 \\
\bottomrule
\end{tabular}%
\caption{Results of different agents at training epoch $= \{100, 200, 300\}$. Each number is averaged over $3$ runs, and each run is tested on $50$ dialogues. (Success: success rate) Switch-DDQ outperforms DQN and DDQ variants after Epoch 100, where DQN(5) is shown as the upper bound as it uses more real experiences. Best scores are labeled in blue.
}
\label{tab:main_tab}%
\end{table*}%

\begin{figure*}[ht!]
\centering
\begin{minipage}[t]{0.47\textwidth}
\centering
\includegraphics[width=1.0\textwidth]{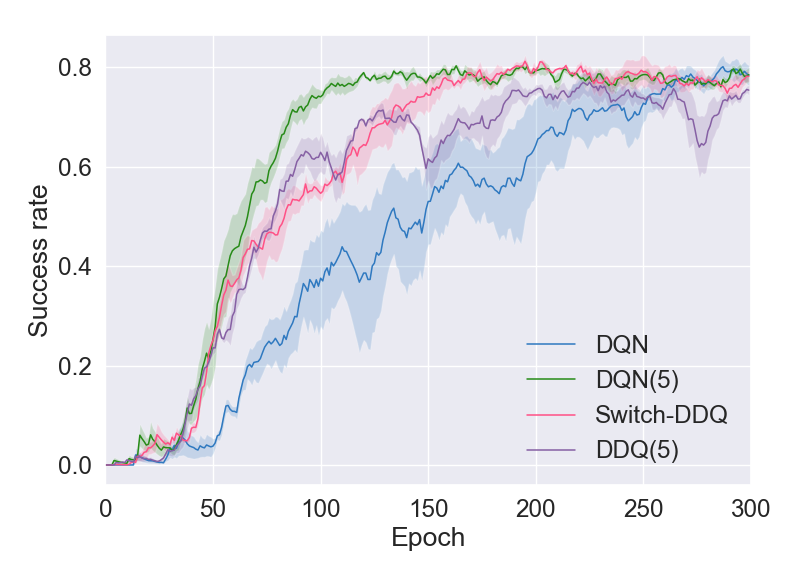}
\caption{The learning curves of DQN, DQN($5$), Switch-DDQ, and DDQ($5$) of each epoch.}
\label{fig:main_epoch}
\end{minipage}%
~\hfill
\begin{minipage}[t]{0.47\textwidth}
\centering
\includegraphics[width=1.0\textwidth]{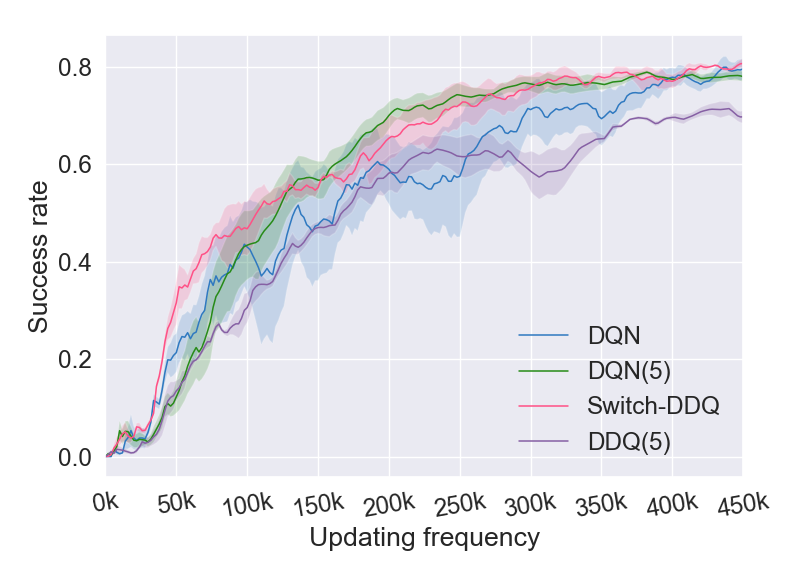}
\caption{The learning curves of DQN, DQN($5$), Switch-DDQ, and DDQ($5$) on the scale of updating frequency. 
}
\label{fig:main_freq}
\end{minipage}
\end{figure*}

\subsection{Simulation Evaluation}
We train the dialogue agents by simulating interactions between the agents and well-programmed user simulators, instead of real users. That is, we train the world model to imitate the behaviors of the user simulator. 
\subsubsection{User Simulator}
We used an open-sourced task-oriented user simulator \cite{li2016user} in our simulated evaluation. At each dialogue turn, the simulator will emit a simulated user response to the agent. When the dialogue ends, a reward signal will be provided. The dialogue is considered successful, if and only if a movie ticket is booked successfully and the information provided by the agent conform to all the constraint slots in the sampled user goal. Each completed dialogue shows either a positive reward $2 * L$ for success, or a negative reward $-L$ for failure, where $L$ is the maximum number of turns in each dialogue and is set to $40$ in our experiments. Furthermore, in each turn, a negative reward $-1$ is provided to encourage shorter dialogue.



\subsubsection{Main Results}
We summarize the main results in Table~\ref{tab:main_tab} and plot the learning curves in Figure~\ref{fig:main_epoch}. As illustrated in Figure~\ref{fig:DDQ_perf}, DDQ($K$) is highly susceptible to parameter $K$. Therefore, we only keep the best performing DDQ($5$) as the baseline in the following figures. DQN($5$), which uses $4$ times more real user experiences to this end, is the upper bound for the corresponding DDQ($5$) method. In Table~\ref{tab:main_tab}, we report success rate, average reward and average number of turns over $3$ different runs for each agent. As is shown, the agent of Switch-DDQ after the first 100 epochs, consistently achieves higher success rates with a smaller number of interaction turns. Again, DDQ(10) and DDQ(20) quickly deteriorate through the training process. In Figure~\ref{fig:main_epoch}, we can observe that in the first $130$ epochs, DDQ($5$) performs slightly better than Switch-DDQ. However, after that, Switch-DDQ surpasses DDQ($5$) and achieves better performance. It only takes Switch-DDQ $180$ epochs to achieve comparable results to DQN($5$), which utilizes 4 times more real experiences, and DDQ($5$) fails to reach similar performance within $300$ epochs. This is expected, as the aggressive simulator sampling policy adopted by DDQ($5$), though helping update the policy network more rapidly in the early stage of training, hurts the performance due to the use of low-quality training instances in the later stage. Note that except for DQN($5$), all the agents are trained using the same number of real experiences in each epoch, differing only the amounts of simulated experiences used (for planning) and how these simulated experiences are generated (via active learning or not). The result show that Switch-DDQ can utilize simulators in a more effective and robust way than DDQ.

\begin{figure*}[ht!]
\centering
\begin{minipage}[t]{0.47\textwidth}
\centering
\includegraphics[width=1.0\textwidth]{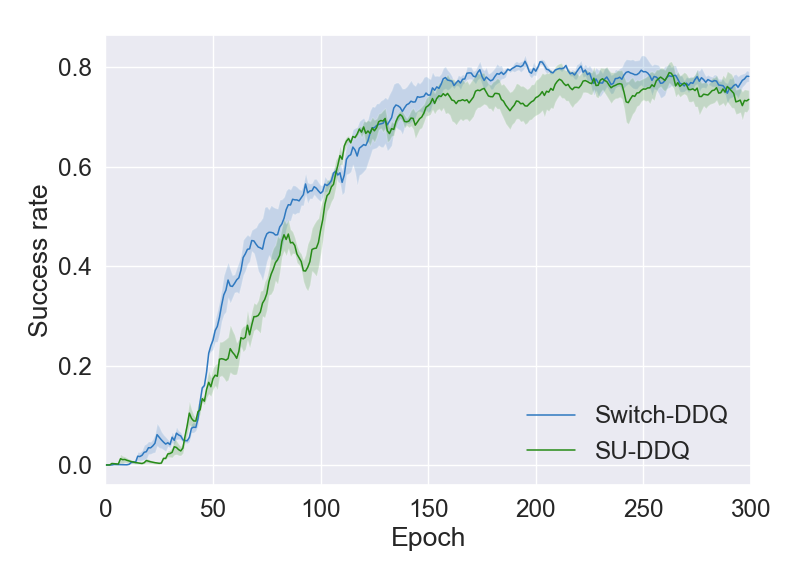}
\caption{Learning curves of Switch-DDQ versus SU-DDQ where SU-DDQ uses a uniform sampling strategy.
}
\label{fig:ablation_epoch}
\end{minipage}%
~\hfill
\begin{minipage}[t]{0.47\textwidth}
\centering
\includegraphics[width=1.0\textwidth]{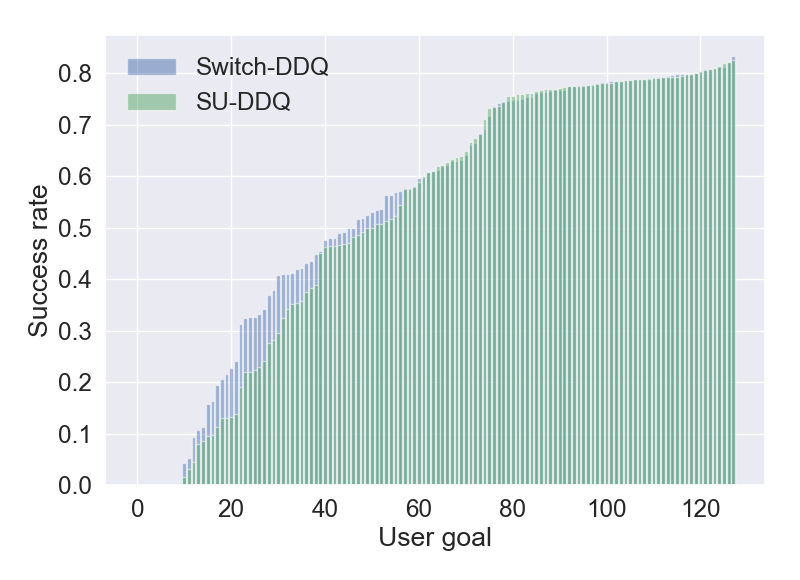}
\caption{Success rate on 128 user goal categories for Switch-DDQ and SU-DDQ, ranking in ascending order.}
\label{fig:ablation_user_goal} 
\end{minipage}
\end{figure*}


We also examine the performance of different agents with an equal number of optimization operations. As shown in Figure~\ref{fig:main_freq}, we plot the success rate as a function of updating frequency, i.e., how many dialogue experiences (either real or simulated) are used altogether to optimize the agent policy network. Note that DQN($5$) displays superior performance over DQN as it generates more diverse dialogues at the same updating frequency (DQN may refer to identical experiences more frequently since $B^u$ in DQN is refreshed less often than that in DQN(5)). 
 Furthermore, we observe that DDQ($5$) fails to obtain a similar performance to DQN, due to the use of many low-quality simulated experiences. However, this does not happen in Switch-DDQ, since it actively samples user goals by making diversified training dialogues and discreetly controlling the amount of simulated experiences via the  switcher.  

\subsubsection{Ablation Test}
To further examine the effectiveness of the active learning module, we conduct an ablation test by replacing the user goal selection routine (Algorithm~\ref{alg:active}) with the one based on uniform sampling, referred to as SU-DDQ. The results in Figure~\ref{fig:ablation_epoch} demonstrate that Switch-DDQ can consistently outperform SU-DDQ, especially in the early phase (before epoch $100$). This is due to the fact that the agent is more sensitive to the diversity of user goals in the earlier stage since in the limited data setting, many repeated cases introduce biases more easily. In Figure~\ref{fig:ablation_user_goal}, we report the success rate for different categories of user goals and rank them in the increasing order. It is observed that for the corresponding rank of user goal category, especially the ones with low success rate, the active version of Switch-DDQ always give a better score. These results demonstrate that the use of the active module improves training efficiency.

\subsection{Human Evaluation}
Real users were recruited to interact with different agents, while the identity of the agent system is hidden from the users. At the beginning of the dialogue session, the user was provided with a randomly sampled user goal, and one of the agents was randomly picked to converse with the user. The dialogue session can be terminated at any time, if the user finds that the dialogue takes so many turns that it is unlikely to reach a promising outcome. Such dialogues are considered as failed in our experiments.

\begin{figure}[ht!]
\centering
\includegraphics[width=0.85\linewidth]{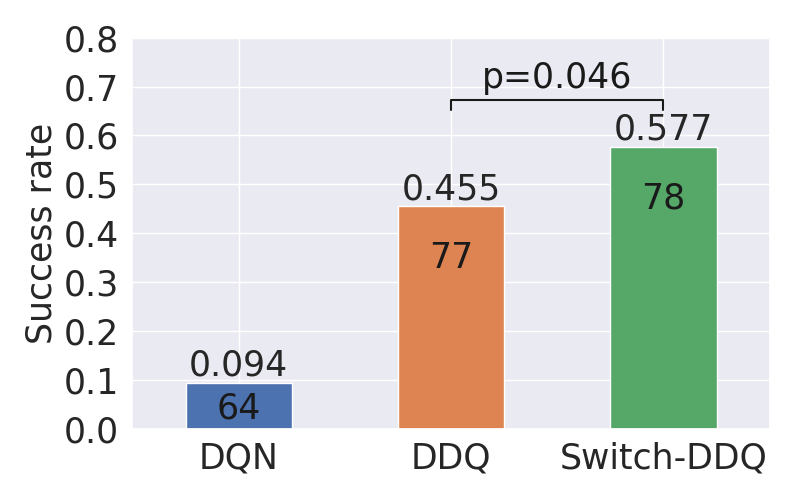}
\caption{Human evaluation results of DQN, DDQ(5), and Switch-DDQ. The number of test dialogues is shown on each bar, and the one-sided p-value is from a two-sample permutation test over the success/fail lists.}
\label{fig:human_eval}
\end{figure}

Three agents (DQN, DDQ(5), and Switch-DDQ) trained as previously described (Figure~\ref{fig:main_epoch}) at epoch 150 are selected as for human evaluation.\footnote{Epoch 150 is picked since we are testing the effectiveness of methods using a small number of real experiences.} As illustrated in Figure~\ref{fig:human_eval}, the results of human evaluation are consistent with those in the simulation evaluation. We find that DQN is abandoned more often by users as it takes so many dialogue turns (Table~\ref{tab:main_tab}) resulting in a much hefty performance drop, and the proposed Switch-DDQ outperforms all the other agents.

\section{Conclusion}
This paper presents a new framework Switch-based Active Deep Dyna-Q (Switch-DDQ) for task-completion dialogue policy learning. With the introduction of a switcher, Switch-DDQ is capable of adaptively choosing the proper data source to use, either from real users or world model, enhancing the efficiency and robustness of dialogue policy learning. Furthermore, the active user goal sampling strategy provides a better utilization of the world model than that of previous DDQ, and boosts the performance of training. Validating Switch-DDQ in the movie-ticket booking task with simulation experiments and human evaluation, we show that the Switch-DDQ agent outperforms the agents trained by other state-of-the-art methods, including DQN and DDQ. Switch-DDQ can be viewed as a generic model-based RL approach, and is easily extensible to other RL problems.

\section*{Acknowledgement}
We thank the reviewers for their helpful comments, and we would like to acknowledge the volunteers for helping us with the human experiments. This work was done in part when Yuexin Wu was visiting Microsoft Research as an intern, and is supported in part by the National Science Foundation (NSF) under grant IIS-1546329.
	
\bibliographystyle{aaai}
\bibliography{saddq}
	
\end{document}